\documentclass[10pt,twocolumn,letterpaper]{article}

\usepackage[pagenumbers]{cvpr} 

\usepackage{xspace}
\usepackage{pifont} 
\usepackage{tikz} 
\usepackage{xcolor} 
\usepackage{multirow}

\usepackage{tocloft}

\newcommand{\ourmethod}{{{SuperHead}}\xspace}
\newcommand\mypara[1]{\vspace{1mm}\noindent\textbf{#1}}

\definecolor{cvprblue}{rgb}{0.21,0.49,0.74}
\usepackage[pagebackref,breaklinks,colorlinks,citecolor=cvprblue]{hyperref}

\usepackage[table]{xcolor}
\definecolor{willy}{rgb}{0.6, 0, 0}

\def\paperID{377} 
\def\confName{3DV\xspace}
\def\confYear{2026\xspace}

\title{From Blurry to Believable:\\Enhancing Low-quality Talking Heads with 3D Generative Priors}

\author{Ding-Jiun Huang$^1$\textcolor{white}{\thanks{Equal Contribution}} \quad Yuanhao Wang$^{1}$ \quad 
Shao-Ji Yuan$^{1}$ \quad Albert Mosella-Montoro$^{1}$\\
Francisco Vicente Carrasco$^{1}$ \quad Cheng Zhang$^{2}$ \quad Fernando De la Torre$^{1}$\\ [6pt]
\textsuperscript{1} Carnegie Mellon University \quad \textsuperscript{2} Texas A\&M University\\ [6pt]
\href{https:/humansensinglab.github.io/super-head/}{humansensinglab.github.io/super-head}
}

\begin{document}
\twocolumn[{%
\renewcommand\twocolumn[1][]{#1}%
\maketitle
\vspace{-2mm}
\centerline{\includegraphics[width=0.98\linewidth]{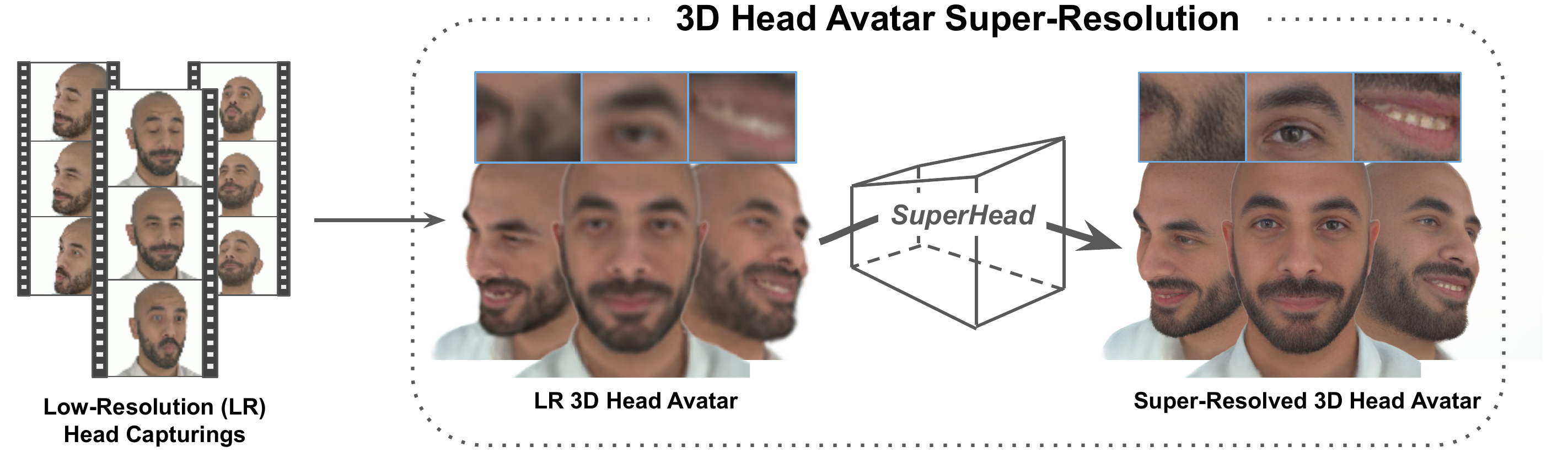}}
    \vspace{-4mm}
    \captionof{figure}{\small We present SuperHead, a new method for super-resolving low-resolution 3D head avatars. Given a low-resolution animatable avatar reconstructed from low-quality captures, SuperHead synthesizes high-fidelity geometry and detailed textures while ensuring multi-view and temporal consistency under diverse facial expressions. Unlike prior image- or video-based SR approaches, our method directly upsamples 3D avatars, enabling photorealistic animation and faithful identity preservation from degraded inputs.}
    \label{fig:teaser}
    \vspace{5mm}
}]
\begin{abstract}
Creating high-fidelity, animatable 3D talking heads is crucial for immersive applications, yet often hindered by the prevalence of low-quality image or video sources, which yield poor 3D reconstructions. In this paper, we introduce \ourmethod, a novel framework for enhancing low-resolution, animatable 3D head avatars. The core challenge lies in synthesizing high-quality geometry and textures, while ensuring both 3D and temporal consistency during animation and preserving subject identity. Despite recent progress in image, video and 3D-based super-resolution (SR), existing SR techniques are ill-equipped to handle dynamic 3D inputs. To address this, \ourmethod leverages the rich priors from pre-trained 3D generative models via a novel dynamics-aware 3D inversion scheme. This process optimizes the latent representation of the generative model to produce a super-resolved 3D Gaussian Splatting (3DGS) head model, which is subsequently rigged to an underlying parametric head model (e.g., FLAME) for animation. The inversion is jointly supervised using a sparse collection of upscaled 2D face renderings and corresponding depth maps, captured from diverse facial expressions and camera viewpoints, to ensure realism under dynamic facial motions. Experiments demonstrate that \ourmethod generates avatars with fine-grained facial details under dynamic motions, significantly outperforming baseline methods in visual quality. 

\end{abstract}    

\section{Introduction}

\begin{figure*}[h]
    \centering
    \vspace{-4mm}    \includegraphics[width=0.98\textwidth]{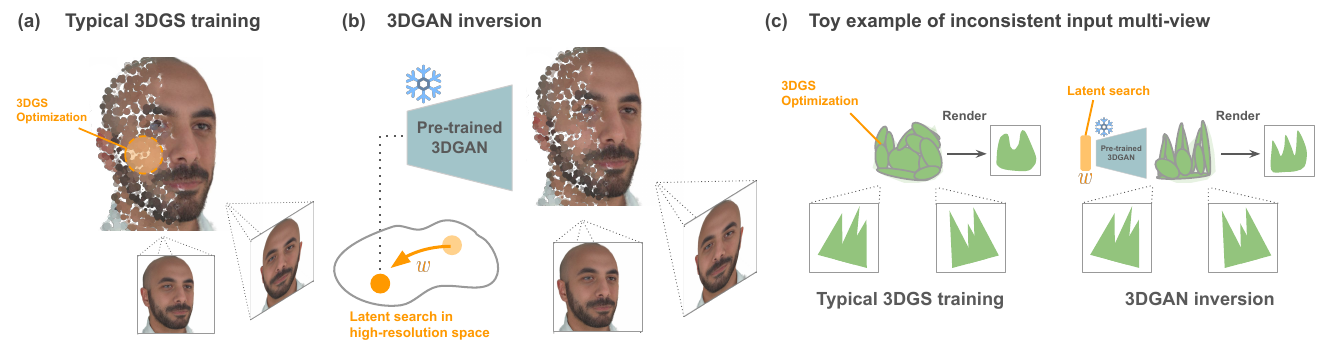}
    \vspace{-5mm}
    \caption{ (a) Typical 3DGS process optimizes a head from multi-view images. (b) 3D GAN inversion aims to find a latent code whose generated 3D head best explains the given multi-view images. (c) When input multi-views are inconsistent, typical 3DGS tends to create an ``averaged" result as a compromise among inconsistent inputs. 3D GAN inversion can generate high-frequency details because it searches in a pre-trained high-resolution space.
    }
    \vspace{-4mm}
    \label{fig:motivation}
\end{figure*}

High-fidelity, animatable 3D head avatars are increasingly important in augmented and virtual reality, telepresence, gaming, and digital entertainment, where realism is critical. However, capturing scan-quality geometry and dynamic textures with traditional pipelines requires costly, specialized equipment~\cite{bickel2007multi,beeler2010high,bradley2010high,ghosh2011multiview}, limiting its accessibility. As a result, much recent work reconstructs dynamic 3D heads from videos or images acquired with consumer devices~\cite{hong2021headnerf,gafni2021dynamic,grassal2021neural,xiao2022detailed,zheng2022imavatar,zheng2023neuface,qian2024gaussianavatars,shao2024splattingavatar,xiang2024flashavatar,kirschstein2024gghead,he2025lam,chu2024gpavatar}. Yet such inputs often suffer from low resolution, poor lighting, and motion blur, producing avatars with blurry textures and artifacts that fall short of high-fidelity requirements.

In this paper, we focus on enhancing the visual quality of low-resolution animatable head avatars, i.e., 3D talking head models whose textures appear blurry and lack fine facial details, often as a result of reconstruction from low-quality inputs. While significant progress has been made in image, video, and static 3D super-resolution (SR), extending these advances to dynamic 3D heads presents unique challenges. Unlike static SR, this task demands simultaneously generating high-resolution geometry and textures, preserving temporal consistency, and faithfully maintaining identity. Existing 3D SR approaches~\cite{wang2022nerf,han2024super,huang2023refsr,yoon2023cross,shen2024supergaussian,liu20243dgs} often rely on 2D priors, but applying them frame-by-frame leads to flickering, and video SR cannot ensure cross-view consistency, making them ill-suited for dynamic avatars. 

Recent advances in 3D-aware generative models, particularly 3D GANs~\cite{yin20233d,yuan2023make,bhattarai2024triplanenet}, have demonstrated strong 3D priors for reconstructing head avatars via GAN inversion. By projecting input images into the latent space of a 3D GAN, they can synthesize high-quality 3D heads from limited data. Inspired by this paradigm, we propose to extend 3D GAN inversion to the super-resolution of animatable 3D avatars. As shown in Figure~\ref{fig:motivation}, we demonstarte the difference between typical 3DGS head reconstruction and 3D GAN inversion. We also provide an example to illustrate that 3D GAN inversion can well handle inconsistent multi-view inputs. However, unlike static inversion, directly applying inversion to animatable avatars is much harder: the super-resolved head must stay realistic and identity-preserving across diverse expressions.

To address these challenges, we propose a dynamics-aware 3D inversion framework for upsampling low-resolution animatable head avatars. The key idea is to leverage the strong 3D priors of generative head models while tightly coupling the inversion process with the underlying mesh structure (e.g., FLAME~\cite{li2017learning}). Instead of relying on single-frame information, we condition inversion on multi-view renderings and depth cues, ensuring cross-view consistency and faithful alignment. To further tackle animation, we extend this conditioning across multiple facial expressions, enabling the reconstructed avatar to preserve identity, detail, and realism even under dynamic motion.

We name our approach \textbf{\ourmethod} and demonstrate its effectiveness on different benchmarks (i.e., NeRSemble~\cite{kirschstein2023nersemble}, INSTA~\cite{zielonka2023instant}). \ourmethod works seamlessly across different head avatar models (e.g., GaussianAvatar~\cite{qian2024gaussianavatars}, SplattingAvatar~\cite{shao2024splattingavatar}), showing clear gains in fidelity over competing methods. Our dynamics-aware design ensures that the upsampled avatars remain consistent under diverse facial motions, a scenario where existing methods often struggle. Moreover, since GAN inversion is computationally lightweight, our method achieves these improvements without sacrificing efficiency, enabling fast inference. To the best of our knowledge, this is the first framework that directly addresses the task of super-resolving low-resolution animatable 3D head avatars and obtains competitive results.

\section{Related Work}

\textbf{Animatable 3D Head Avatar.}
3D head avatar modeling has long been an active area due to its broad applications. Early works rely on multi-view stereo for geometry reconstruction~\cite{bickel2007multi,beeler2010high,bradley2010high,ghosh2011multiview}, but require heavy computation and complex capture setups. More recent approaches employ volumetric representations~\cite{hong2021headnerf,yao2022dfa,guo2021adnerf,zhuang2022mofanerf,sun2022fenerf,athar2022rignerf,gafni2021dynamic,grassal2021neural} or neural implicit functions~\cite{yenamandra2020i3dmm,xiao2022detailed,zheng2022imavatar,zheng2022imface,zheng2023neuface}, offering higher efficiency and reconstruction quality, often leveraging 3D Morphable Models~\cite{blanz2003face,li2017learning} for semantic control and identity–expression disentanglement. INSTA~\cite{zielonka2023instant} deforms query points to a canonical space using FLAME \cite{li2017learning}, while Khakhulin et al.~\cite{khakhulin2022realistic} learn non-rigid deformations on FLAME templates to recover dynamic details from in-the-wild videos. Another branch of methods \cite{chu2024gpavatar,zheng2024headgap} explore 3D head avatar creation with few-shot images, avoiding the need of heavy video inputs. 
More recently, 3D Gaussian Splatting (3DGS) \cite{kerbl3Dgaussians} has advanced novel view synthesis with superior fidelity and much faster rendering, and has become widely used for head avatar modeling \cite{qian2024gaussianavatars,shao2024splattingavatar,xiang2024flashavatar,liao2023hhavatar,zheng2024headgap,lee2025surfhead,xu2024gphmv2,kirschstein2024gghead,chu2024gagavatar,zhou2024headstudio,dhamo2024headgas}. GaussianAvatars \cite{qian2024gaussianavatars} binds 3D Gaussians to FLAME mesh surface to create animatable heads; SplattingAvatar \cite{shao2024splattingavatar} offsets Gaussians along surface normals to fit more complex geometry; and SurFhead \cite{lee2025surfhead} rigs 2D Gaussian surfels to parametric models for well-defined geometry. Our method is complementary to these designs and can be directly applied to different types of Gaussian-based head avatars to enhance their fidelity and robustness.

\begin{figure*}[t]
    \centering
    \includegraphics[width=1\linewidth]{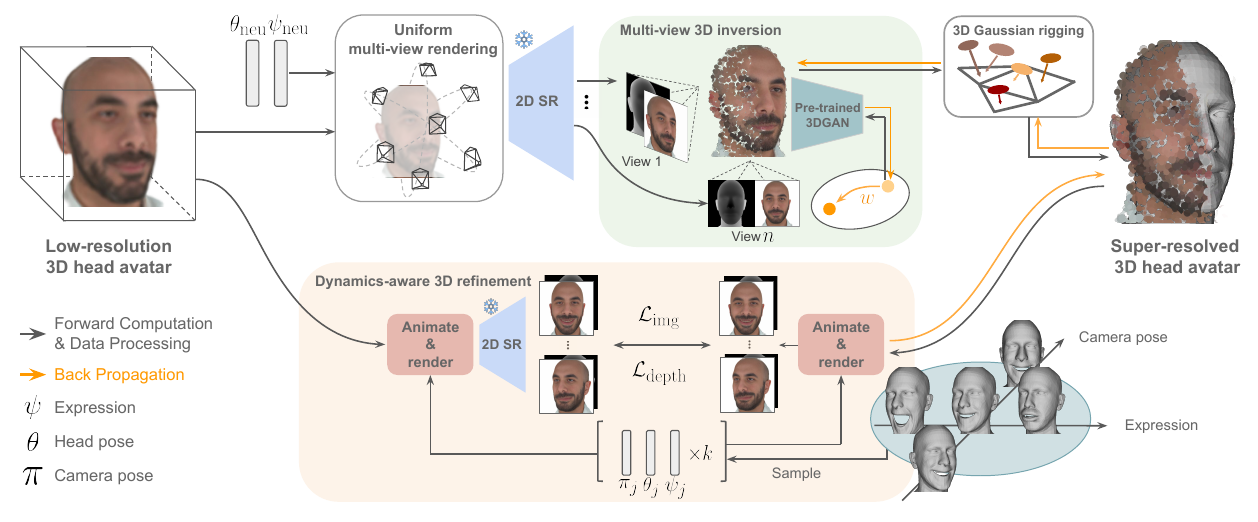}
    \vspace{-5mm}
    \caption{\textbf{Overview of \ourmethod.} Given a low-resolution 3D head avatar driven by a morphable model, we first reconstruct static 3D head in the canonical space with multi-view 3D GAN inversion (Section~\ref{multi_view_3d_inversion}). We then refine mesh geometry and rig 3D Gaussians onto mesh surface to enable animation (Section~\ref{3d_gaussian_rigging}).
    We further include anchor images with diverse camera poses and expressions for dynamics-aware 3D refinement, ensuring the robustness of the 3D head model across viewing angles and complex facial motions (Section~\ref{dynamics_aware_inversion}).}
    \vspace{-3mm}
    \label{fig:framework}
\end{figure*}

\mypara{3D Super-Resolution.}
Despite progress in 3D reconstruction and generation, enhancing low-resolution 3D content remains challenging. One line of work~\cite{han2024super,huang2023refsr,lee2024disr,wang2022nerf,yoon2023cross} leverages pre-trained single-image super-resolution (SISR) models to upscale images rendered from 3D representations, with additional mechanisms to improve multi-view consistency. More recently, other methods~\cite{liu20243dgs,shen2024supergaussian,ko2025sequence} employ pre-trained video upsamplers~\cite{xu2024videogigagan,blattmann2023stable} to enhance 3D content through 2D video priors, offering stronger temporal coherence but struggling with long-range consistency under large viewpoint changes. In contrast, our approach achieves dynamic 3D super-resolution by directly enforcing both multi-view and temporal consistency, producing high-fidelity avatars robust to diverse expressions and motions.

\mypara{Generative Head Avatar Reconstruction.}
Advances in 3D-aware generative models~\cite{chan2022efficient,an2023panohead,gu2021stylenerf,sun2023next3d,kirschstein2024gghead,hyun2024gsgan} have gained attention for synthesizing high-quality 3D representations and consistent multi-view images, making them strong priors for head avatar reconstruction. Among them, GSGAN~\cite{hyun2024gsgan} generates 3D Gaussian-based static head through adversarial training with large-scale 2D head image data. Several works~\cite{bhattarai2024triplanenet,fruhstuck2023vive3d,ko20233d,trevithick2023real,xu2023n,yin20233d,yuan2023make} reconstruct and edit 3D faces from a single image via GAN inversion, projecting real images into the latent space of 3D-aware GANs to enable viewpoint-consistent synthesis and semantic control. Others target animatable avatars. AniFaceGAN \cite{wu2022anifacegan} extends a static 3D-aware generator \cite{deng2022gram} with a deformation field mapping query points to canonical space, while Lin et al.~\cite{lin20223d} use 3D GAN inversion to animate portraits. Next3D~\cite{sun2023next3d} combines a tri-plane representation with a conditional 3D Morphable Model (3DMM) and applies PTI inversion~\cite{roich2022pivotal}, while InvertAvatar~\cite{zhao2024invertavatar} improves reconstruction through incremental inversion across multiple frames. These works show the effectiveness of pre-trained 3D GANs for one-shot or few-shot avatar reconstruction. We extend this direction by leveraging 3D generative priors to upsample low-resolution animatable head avatars.

\section{Preliminary}\label{Preliminary}
We use 3D GAN inversion with GSGAN~\cite{hyun2024gsgan} for 3D head reconstruction and adopt a representation where avatars are modeled as 3D Gaussian splats rigged to a parametric face model. In this section, we review both components.
\subsection{3D GAN Inversion}\label{pre:gan_inversion}
Given a pre-trained unconditional 3D GAN model $G$ parameterized by weight $g$ and an input image $\mathbf{I}$, the goal of 3D GAN inversion is to find a latent code $\boldsymbol{w}^{*}$ that accurately reconstructs $\mathbf{I}$ at the corresponding camera pose, while enabling content-consistent synthesis from novel viewpoints. This problem is commonly formulated as:
\begin{equation}
    \boldsymbol{w}^* = \arg\min_{\boldsymbol{w}} \mathcal{L}_\text{recon}( G (\boldsymbol{w}, \pi; g), \mathbf{I}),
\end{equation}
where $\boldsymbol{w}$ is the latent representation in $\mathcal{W}^+$ space~\cite{karras2019style,karras2020analyzing} of the 3D GAN model, $\pi$ is the corresponding camera matrix of the input image, and $\mathcal{L}_\text{recon}$ is a pixel-wise reconstruction or perceptual loss. However, optimizing only in the $\mathcal{W}^+$ space often fails to capture fine-grained facial details, resulting in suboptimal reconstructions. To address this issue, recent methods (e.g.,  \cite{roich2022pivotal}) propose a second-stage inversion in the parameter space by slightly finetuning the GAN generator, and achieves improved reconstruction results.  This second-stage optimization is formulated as:
\begin{equation}
    g^* = \arg\min_{g} \mathcal{L}_\text{recon}({G}(\boldsymbol{w}^*, \pi; g), \mathbf{I})
\end{equation}
where $\boldsymbol{w}^*$ is fixed from the first stage. In our framework, we thus adopt this two-stage 3D GAN inversion approach for high-fidelity reconstruction. In this paper, we define ``anchor image'' as all images $\mathbf{I}$ that are used for inversion.

\subsection{3D Gaussian Avatar}
Following GaussianAvatars~\cite{qian2024gaussianavatars}, an avatar is represented by 3D Gaussian primitives anchored to the surface of a parametric face model such as FLAME~\cite{li2017learning}. Each primitive is defined in the local coordinate system of a mesh face by position $\boldsymbol{\mu}'$, rotation $\mathbf{r}'$, and scaling $\mathbf{S}'$, and transformed to global coordinates as:
\begin{equation} \label{gaussian_local2global}
    \mathbf{r} = \mathbf{R}\mathbf{r}', \quad \boldsymbol{\mu} = s\mathbf{R} \boldsymbol{\mu}' + \mathbf{T}, \quad \mathbf{S} = s \mathbf{S}',
\end{equation}
where $\mathbf{R}$ is the orientation of the triangle face, $\mathbf{T}$ is the mean of its three vertices, and $s$ is a scale factor. The FLAME model is controlled by shape parameters $\beta$, pose parameters $\theta$, expression parameters $\psi$, and static vertex offsets $\delta$.

\section{Our Approach}
\textbf{Overview.} Figure~\ref{fig:framework} illustrates the \ourmethod framework. The input is a low-resolution (LR) animatable 3D head model $H$, whose movement is driven by an underlying FLAME model $M$. Our goal is to synthesize an animatable head avatar $\hat{H}$ that exhibits a high level of realism and fidelity in various facial expressions, while preserving the identity of the subject. We start by synthesizing a static 3D Gaussian head in canonical space through multi-view 3D GAN inversion (Section~\ref{multi_view_3d_inversion}), where multi-view renderings of a neutral expression $\psi_{\textrm{neu}}$ from LR input model are sampled and super-resolved for the inversion. Then, we rig the static 3D Gaussian head to the FLAME mesh for animation (Section~\ref{3d_gaussian_rigging}). Finally, to make the synthesized 3D head model robust under different facial motions, we perform dynamics-aware 3D refinement to get $\hat{H}$ by jointly optimizing 3D GAN inversion using a set of sampled and super-resolved anchor images $\{ \hat{\mathbf{I}}_j \}_{j=1}^k$ (Section~\ref{dynamics_aware_inversion}) rendered from the LR input model $H$ with configurations $\{ (\theta_j, \psi_j, \pi_j) \}_{j=1}^k$, where $\theta_j$, $\psi_j$ are the pose and expression parameters of FLAME and $\pi_j$ is the camera parameter.

\subsection{Multi-View 3D Inversion} \label{multi_view_3d_inversion}

Given a pre-trained 3D GAN $G$ for generating Gaussian head avatars with parameters $g$, we optimize a latent code $\boldsymbol{w}^*$ such that the resulting 3DGS representation $\mathcal{G} = G(\boldsymbol{w}^*, g^*)$ can faithfully reconstruct a given set of $n$ multi-view anchor 
images $\{\hat{\mathbf{I}}_i\}_{i=1}^n$.

\mypara{Multi-view anchor images sampling}. We collect $\{\mathbf{I}_i\}_{i=1}^n$ by sampling $n$ multi-view renderings of a specific expression $\psi_{\textrm{neu}}$ and FLAME mesh pose $\theta_{\textrm{neu}}$ from LR input model. The camera poses $\pi_i$ of the multi-views are uniformly sampled on a sphere with radius $r$ centered at the model. Note that we only sample views within an angle range that covers the frontal part of input head model. We manually select $\psi_{\textrm{neu}}$ with a major consideration that $\psi_{\textrm{neu}}$ should contain major occluded parts, e.g., teeth and eyeballs, so that $G$ can generate these facial parts through the inversion. Then, we obtain high-resolution counterparts $\{\hat{\mathbf{I}}_i\}_{i=1}^n$  using an off-the-shelf image super-resolution model \cite{zhou2022towards}. 

\mypara{Multi-view 3D GAN inversion}. We then reconstruct a static 3D Gaussian head in canonical space using the multi-view images $\{\hat{\mathbf{I}}_i\}_{i=1}^n$ and the canonical FLAME mesh $\hat{M}_{\theta_{\textrm{neu}},\psi_{\textrm{neu}}}$, defined by the neutral pose $\theta_{\textrm{neu}}$ and expression $\psi_{\textrm{neu}}$. The objective is to minimize the reconstruction loss:
\begin{align}\label{img_loss}
\mathcal{L}_{\text{img}}^{i} = \left\| \hat{\mathbf{I}}_i' - \hat{\mathbf{I}}_i \right\|_2 
+ \lambda_\text{p}\,\mathcal{L}_{\text{pips}}(\hat{\mathbf{I}}_i', \hat{\mathbf{I}}_i),
\end{align}
where $\hat{\mathbf{I}}_i' = \mathcal{R}(G(\boldsymbol{w}, g); \pi_i)$ and $\mathcal{R}$ is a differentiable renderer. The object $\mathcal{L}_{\text{pips}}$ is the perceptual loss, and $\pi_i$ is the camera parameter of $\hat{\mathbf{I}}_i$.

To enforce geometric alignment, we add depth supervision. Let $D_i$ be the depth map of $\hat{M}_{\theta_{\textrm{neu}},\psi_{\textrm{neu}}}$ rendered at $\pi_i$, and $m_i$ a mask excluding hair regions. The depth loss is
\begin{align}\label{depth_loss}
\mathcal{L}_{\text{depth}}^{i} = \left\| \big(\mathcal{R}_d(G(\boldsymbol{w}, g); \pi_i) - D_i\big) \cdot m_i \right\|_2 ,
\end{align}
where $\mathcal{R}_d$ is the differentiable depth renderer. Depth supervision is omitted for hair regions since FLAME does not model hair. The multi-view 3D GAN inversion loss is 
\begin{align}\label{loss_mv}
\mathcal{L}_\text{mv} = \frac{1}{n} \sum_{i=1}^n 
\big( \mathcal{L}_\text{img}^i + \lambda_\text{d}\,\mathcal{L}_\text{depth}^i \big).
\end{align}
The inversion jointly enforces image and depth consistency across views, ensuring the reconstructed 3D Gaussian head aligns photometrically and geometrically with the input.

\subsection{3D Gaussian Rigging} \label{3d_gaussian_rigging}

Following GaussianAvatars~\cite{qian2024gaussianavatars}, we bind the reconstructed Gaussian head $\mathcal{G}$ to the underlying FLAME mesh for animation. However, the initial geometry of the low-resolution head model $H$, driven by the FLAME mesh $M$, often deviates from the super-resolved images $\{\hat{\mathbf{I}}_i\}_{i=1}^n$, so directly binding $\mathcal{G}$ to the mesh may cause mismatching between 3D Gaussians and corresponding facial parts. For example, 3D Gaussians of teeth could be bound to lips on the mesh. To correct these misalignments, we first perform mesh geometry refinement, then bind 3D Gaussians to the refined mesh. 

\mypara{Geometry refinement}. Let $M_{\beta,\theta,\psi}$ denote the FLAME mesh parameterized by global shape $\beta$, pose $\theta$, and expression $\psi$. For each image $\hat{\mathbf{I}}_{i \leq n}$, we detect 2D facial landmarks $\ell_i \in \mathbb{R}^{2\times m}$, where $m$ is the number of landmarks. The corresponding 3D landmarks are extracted as $\mathbf{X}_i = S(M_{\beta,\theta_i,\psi_i}) \in \mathbb{R}^{3\times m}$, where $S(\cdot)$ is a landmark selection operator that selects the $m$ predefined mesh vertices corresponding to facial landmarks. With camera projection $\Pi_i(\cdot)$, the optimized shape parameter is 
\begin{align}
\beta^* = \arg\min_\beta \sum_{i=1}^n 
\left\| \Pi_i(\mathbf{X}_i) - \ell_i \right\|_2.
\end{align}
Here, $\beta$ is shared globally across all anchor images, while $(\theta_i,\psi_i,\Pi_i)$ vary per image. The optimized shape $\beta^*$ is fixed for the rest of the pipeline. 
See Figure~\ref{fig:flame_refinement_ablation} for illustrations.

\mypara{3D Gaussian rigging}. Following GaussianAvatars~\cite{qian2024gaussianavatars}, we bind the reconstructed Gaussian head $\mathcal{G}$ to the underlying FLAME mesh. Each Gaussian primitive is associated with its nearest mesh face. To enable animation, the Gaussian parameters (i.e., position $\boldsymbol{\mu}$, rotation $\mathbf{r}$, and scale $\mathbf{S}$) are transformed into the local coordinate system of that face as
\begin{align}
\mathbf{r}' = \mathbf{R}^{-1}\mathbf{r}, 
\boldsymbol{\mu}' = \frac{1}{s}\,\mathbf{R}^{-1}(\boldsymbol{\mu} - \mathbf{T}), 
\mathbf{S}' = \frac{1}{s}\,\mathbf{S}^{-1},
\end{align}
where $\mathbf{R}$ is the face orientation, $\mathbf{T}$ is the mean of its three vertices, and $s$ is a scaling factor. After the FLAME mesh deforms (e.g., under expression changes), the local Gaussian parameters are mapped back to the global coordinate system using Equation~\ref{gaussian_local2global}.

\subsection{Dynamics-Aware 3D GAN Refinement}\label{dynamics_aware_inversion}

Through multi-view 3D GAN inversion (Section~\ref{multi_view_3d_inversion}) and 3D Gaussian rigging (Section~\ref{3d_gaussian_rigging}), we can animate a high-resolution 3D head model. However, this setup has the following limitations: First, there is no guarantee of faithful reconstruction at camera poses far from $\pi_{i \leq n}$; Second, some facial regions are occluded in the canonical view (e.g., eyelids, inner mouth), causing incomplete reconstruction; Finally, $\mathcal{G}$ is unstable under deformation, causing artifacts in large facial motions. To solve the above limitations, we utilize multi-expression renderings from LR input model to jointly optimize the 3D GAN inversion process. 

\begin{figure}[t]
    \centering
    \includegraphics[width=1\linewidth]{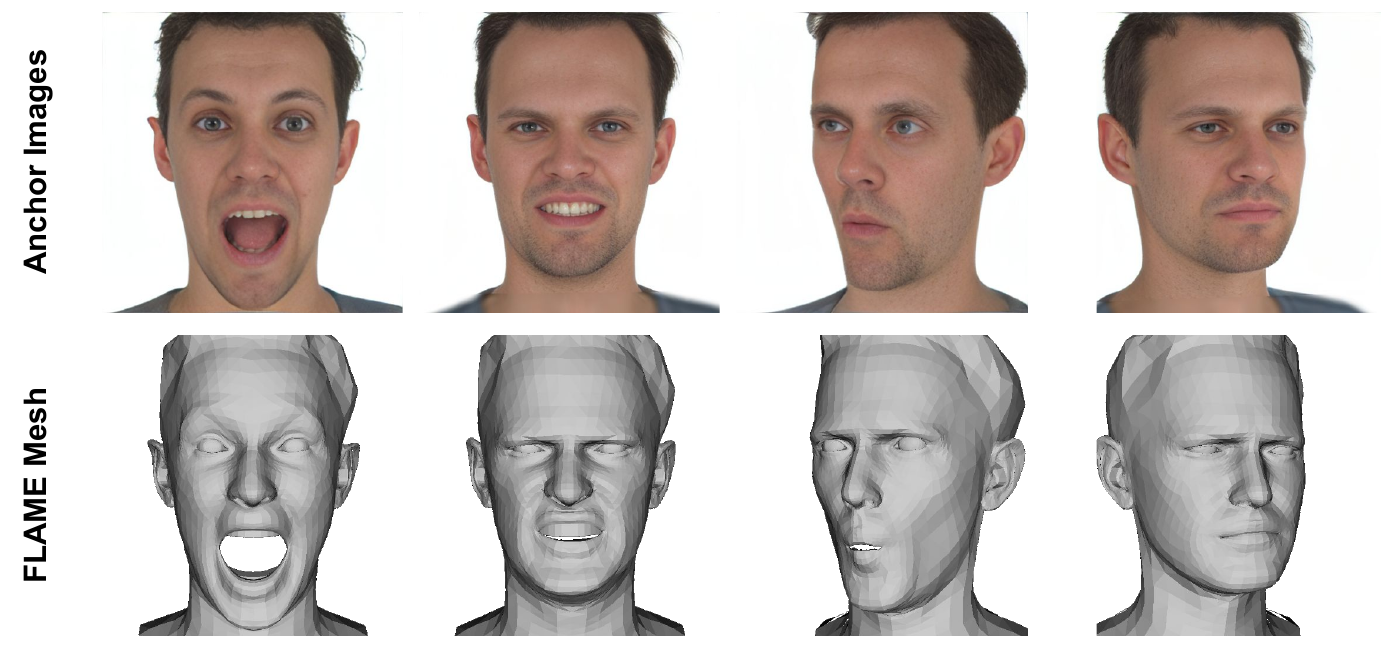}
    \vspace{-5mm}
    \caption{Example of sampled anchor images and the corresponding FLAME mesh used for 3D GAN inversion.
    }
    \vspace{-3mm}
    \label{fig:anchor}
\end{figure}

\mypara{Multi-expression anchor images sampling}. To begin, we first augment the anchor image set $\{\hat{\mathbf{I}}_i\}_{i=1}^n$ by adding super-resolved multi-expression anchor images, getting an augmented image set of $k$ images in total. We manually select various facial expressions that (1) each facial region (e.g., teeth, eyelids) is visible in at least one selected expression, and (2) the selected expressions capture a wide range of common facial deformations to improve robustness under stretch and compression. With the selected expressions, we sample and super-resolve multi-view renderings from LR input model following the same way in Section~\ref{multi_view_3d_inversion}. Note that in the augmented image set, $\{\hat{\mathbf{I}}_i\}_{i=1}^n$ is solely used in multi-view 3D GAN inversion, while $\{\hat{\mathbf{I}}_j\}_{j=1}^k, k>n$ is used for dynamics-aware 3D GAN refinement. See Figure~\ref{fig:anchor} for representative examples. 

\begin{table*}[t]
\centering
\tabcolsep 3pt
\caption{State-of-the-art comparison of upsampling animatable 3D head avatars on the NeRSemble and INSTA datasets. Image metrics (PSNR$\uparrow$, SSIM$\uparrow$, LPIPS$\downarrow$) are computed frame-by-frame using an identical validation motion sequence on novel camera poses and expressions. We also report inference time for each methods, showing that our method achieves best quality with competitive efficiency. 
}
\label{tab:main_results}
\small
\vspace{-3mm}
\resizebox{2\columnwidth}{!}{
\begin{tabular}{lcccccccccc}
\toprule
\multirow{4}{*}{\textbf{Method}} & \multicolumn{6}{c}{\textbf{NeRSemble dataset}~\cite{kirschstein2023nersemble}} & \multicolumn{3}{c}{\textbf{INSTA dataset}~\cite{zielonka2023instant}} & \multirow{2.5}{*}{\textbf{Time}}\\
\cmidrule(lr){2-7} \cmidrule(lr){8-10}
 & \multicolumn{3}{c}{\textbf{Self-Reenactment}} & \multicolumn{3}{c}{\textbf{Novel View Synthesis}} & \multicolumn{3}{c}{\textbf{Self-Reenactment}}\\
\cmidrule(lr){2-4} \cmidrule(lr){5-7} \cmidrule(lr){8-10} \cmidrule(lr){11-11}
& PSNR $\uparrow$ & SSIM $\uparrow$ & LPIPS $\downarrow$ & PSNR $\uparrow$ & SSIM $\uparrow$ & LPIPS $\downarrow$ & PSNR $\uparrow$ & SSIM $\uparrow$ & LPIPS $\downarrow$ & min \\
\midrule
GaussianAvatars (LR) \cite{qian2024gaussianavatars} & 18.56 & 0.811 & 0.302 & 18.72 & 0.822 & 0.249 & 19.79 & 0.837 & 0.220 & ---\\
Video-based SR \cite{feng2024kalman} & \cellcolor{orange!25}21.91 & \cellcolor{orange!25}0.840 & \cellcolor{yellow!25}0.254 & \cellcolor{yellow!25}21.18 & \cellcolor{yellow!25}0.833 & \cellcolor{yellow!25}0.219 & \cellcolor{orange!25}23.01 & \cellcolor{orange!25}0.850 & \cellcolor{yellow!25}0.158 & 30 \\
SuperGaussian \cite{shen2024supergaussian} & \cellcolor{yellow!25}19.57 & \cellcolor{yellow!25}0.837 & 0.278 & \cellcolor{orange!25}21.57 & \cellcolor{orange!25}0.860 & \cellcolor{orange!25}0.215 & \cellcolor{yellow!25}22.89 & \cellcolor{yellow!25}0.842 & 0.177 & \cellcolor{yellow!25}14 \\
SR + GPAvatar \cite{chu2024gpavatar} & 18.29 & 0.768 & \cellcolor{orange!25}0.244 & 18.14 & 0.810 & 0.228 & 20.15 & 0.823 & \cellcolor{orange!25}0.156 & \cellcolor{red!25}3 \\
\textbf{\ourmethod (ours)} & \cellcolor{red!25}22.21 & \cellcolor{red!25}0.850 & \cellcolor{red!25}0.238 & \cellcolor{red!25}22.76 & \cellcolor{red!25}0.871 & \cellcolor{red!25}0.213 & \cellcolor{red!25}23.76 & \cellcolor{red!25}0.864 & \cellcolor{red!25}0.135 & \cellcolor{orange!25}5 \\
\bottomrule
\end{tabular}%
}
\vspace{-4mm}
\end{table*}


\begin{figure*}[t]
    \centering
    \vspace{1mm}
    \includegraphics[width=1\linewidth]{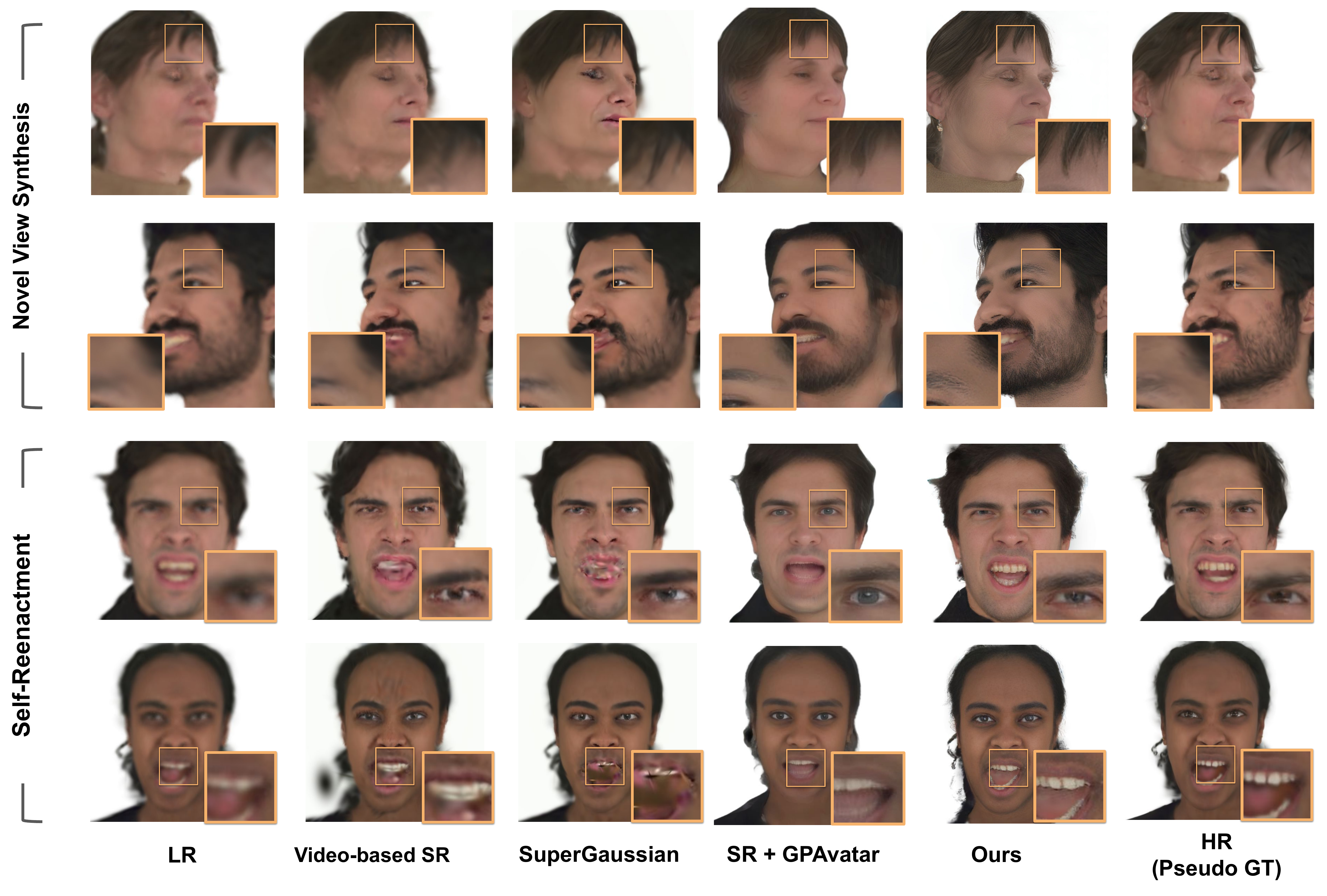}
    \vspace{-9mm}
    \caption{Qualitative comparisons on the NeRSemble dataset~\cite{kirschstein2023nersemble}. \ourmethod synthesizes high-quality facial details across diverse expressions, clearly outperforming baselines and in some cases approaching the pseudo ground-truth head avatar.
    }
    \vspace{-2mm}
    \label{fig:sota-compare}
\end{figure*}

\mypara{Joint optimization via multi-expression anchor images.}
We leverage images with different poses and expressions to jointly supervise 3D GAN inversion. Let $\mathcal{T}_j$ denote the deformation from the canonical mesh $\hat{M}_{\theta_{\textrm{neu}},\psi_{\textrm{neu}}}$ to $\hat{M}_{\theta_j,\psi_j}$. This yields transformed Gaussian heads $\mathcal{T}_j(\mathcal{G})$, enabling both image and depth supervision at every view.
The per-view image reconstruction loss is
\begin{align}
\mathcal{L}_\text{img}^j = 
\left\| \hat{\mathbf{I}}_j' - \hat{\mathbf{I}}_j \right\|_2 
+ \lambda_\text{p}\,\mathcal{L}_{\text{pips}}(\hat{\mathbf{I}}_j', \hat{\mathbf{I}}_j),
\end{align}
where $\hat{\mathbf{I}}_j' = \mathcal{R}(\mathcal{T}_j(G(\boldsymbol{w}, g)); \pi_j)$ is the rendered image at view $\pi_j$.
The per-view depth loss is
\begin{align}
\mathcal{L}_\text{depth}^j = 
\left\| (D_j' - D_j)\cdot m_j \right\|_2,
\end{align}
where $D_j' = \mathcal{R}_d(\mathcal{T}_j(G(\boldsymbol{w}, g)); \pi_j)$ is the rendered depth map and $m_i$ excludes hair regions.
The total objective averages both terms across all views:
\begin{align}\label{loss_dyn}
\mathcal{L}_\text{dyn} = \frac{1}{k} \sum_{j=1}^k 
\big( \mathcal{L}_\text{img}^j + \lambda_\text{d}\,\mathcal{L}_\text{depth}^j \big).
\end{align}
Finally, we can have the final loss term in each optimization iteration by combining Equation~\ref{loss_mv} and Equation~\ref{loss_dyn}:
\begin{align}\label{loss_fn}
\mathcal{L}_\text{total} = \mathcal{L}_\text{mv} + \mathcal{L}_\text{dyn}
\end{align}
Following Section~\ref{pre:gan_inversion} in 3D GAN inversion, we adopt a two-stage training strategy: first optimize $\boldsymbol{w}$ with Equation~\ref{loss_fn}, then fine-tune $g$ with fixed $\boldsymbol{w}^*$.

\section{Experiments} 
We validate \ourmethod with diverse scenarios using different Gaussian avatar models. Section~\ref{setup} describes the experimental setup, Section~\ref{results} presents the main results, and Section~\ref{ablation} provides ablations and analysis. 

\begin{figure*}[t]
    \centering
    \includegraphics[width=1\linewidth]{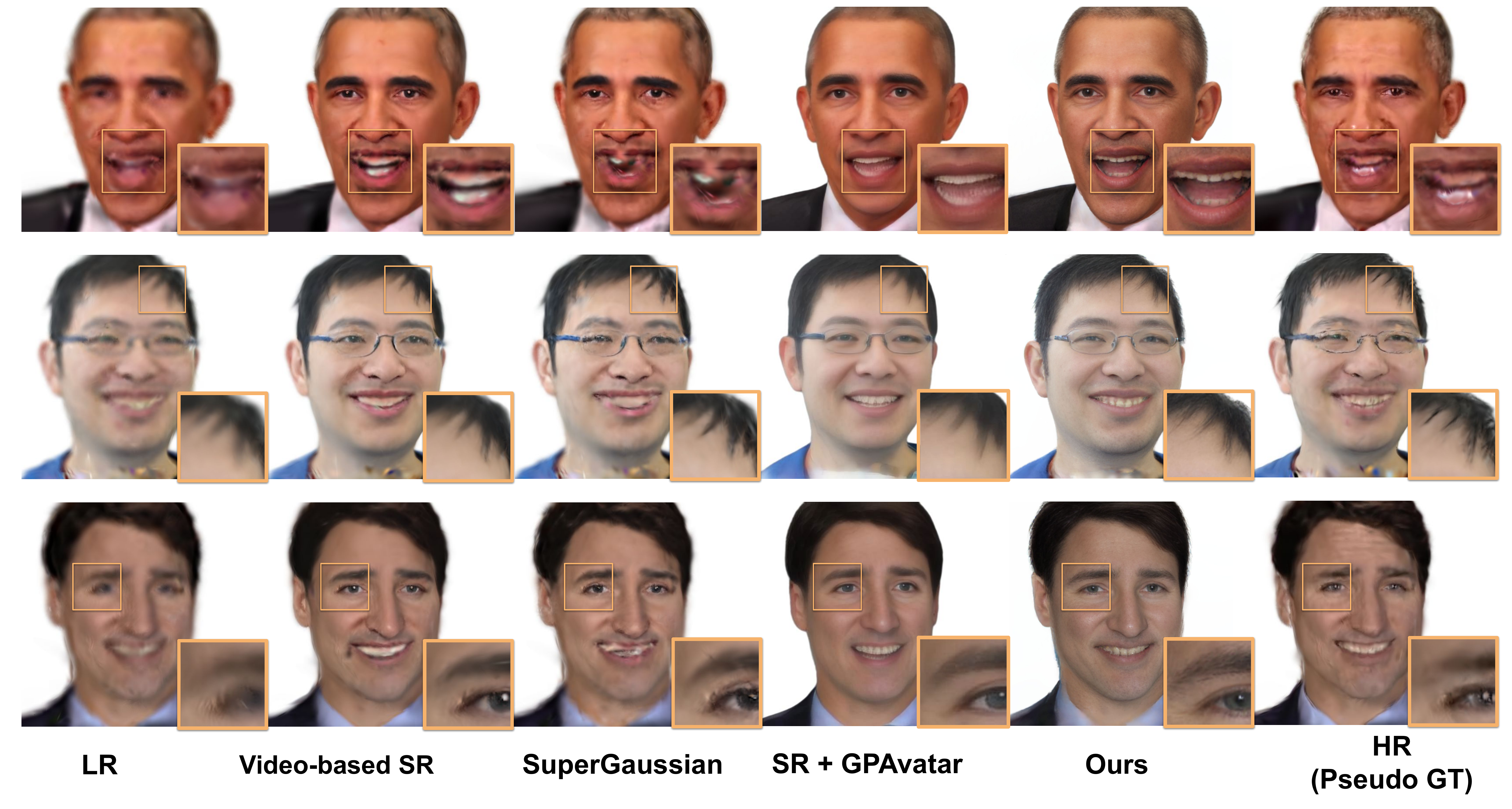}
    \vspace{-8mm}
    \caption{Qualitative results on INSTA dataset~\cite{zielonka2023instant}. All methods are driven and rendered with novel camera poses and expressions.
    }
    \vspace{-4mm}
    \label{fig:insta_qualitative}
\end{figure*}

\subsection{Setup}\label{setup}
\mypara{Datasets.}
We evaluate our method using two challenging datasets: NeRSemble \cite{kirschstein2023nersemble} (multi-view video) and INSTA \cite{zielonka2023instant} (monocular video). For NeRSemble, a pseudo ground-truth animatable head avatar (HR avatar) was reconstructed using GaussianAvatars~\cite{qian2024gaussianavatars} from all multi-view recordings of a single animation sequence at a base resolution of $802 \times 550$. For INSTA, we train the HR avatar with monocular video. We use a down-sampling factor of $4$ to create low-resolution videos. Each set of down-sampled videos was subsequently used to reconstruct a low-resolution head avatar (LR avatar) with GaussianAvatars, which serve as the input to our enhancement pipeline.

\mypara{Evaluation protocols.}
For NeRSemble, we consider two settings, both with baselines driven by the validation sequence: (1) \textit{Novel View Synthesis}, with novel camera poses $\pi_{\textrm{eval}} \notin \pi_{1 \leq j \leq k}$. (2) \textit{Self-Reenactment}, with novel expressions $\psi_{\textrm{eval}} \notin \psi_{1 \leq j \leq k}$. We also show HR avatar as an upper-bound of 3D head super-resolution in qualitative results. For INSTA, which contains only monocular videos, we evaluate in the \textit{Self-Reenactment} setting and report renderings with novel poses and expressions for all methods.

\mypara{Metrics.}
We evaluate the quality of up-scaled animatable 3D head avatars using Peak Signal-to-Noise Ratio (PSNR), Structural Similarity Index (SSIM) \cite{wang2004image}, and Learned Perceptual Image Patch Similarity (LPIPS) \cite{zhang2018unreasonable}.

\mypara{Baselines.}
We compare \ourmethod with the state-of-the-art methods: 
(1)~\textit{GaussianAvatars (LR)}~\cite{qian2024gaussianavatars}: We train it directly on low-resolution videos.
(2)~\textit{Video-based SR}: Given a LR avatar and an animation sequence, we render multi-view video sequences and apply a video SR model~\cite{feng2024kalman} to upsample them independently. GaussianAvatars reconstructs the HR avatar from these upsampled videos.
(3)~\textit{3D-based SR}: A naive baseline for upsampling an animatable 3D model is to upsample the static model in the canonical pose. Following this idea, we apply SuperGaussian~\cite{shen2024supergaussian} to upsample the static 3D Gaussian head under $\psi_{\textrm{neu}}$ expression. The resulting HR head is then rigged to the original FLAME model and animated over the motion sequence.
(4)~\textit{SR+GPAvatar}: We reconstruct a 3D head model using the few-shot method GPAvatar~\cite{chu2024gpavatar}, with super-resolved multi-view renderings from the LR avatar as input.

\mypara{Implementation details.}
We use pre-trained GSGAN~\cite{hyun2024gsgan} for 3D GAN inversion. We place cameras on a sphere of radius $r=0.6$, with $\pi_{1 \leq j \leq k}$ sampled within $\pm 45^\circ$ across all axes and facing the LR input center. Each optimization runs 150 iterations with both multi-view inversion and dynamics-aware refinement. We use Adam with learning rate $0.04$ for $\mathcal{W}^+$ inversion, followed by PTI~\cite{roich2022pivotal} for the last 50 iterations with learning rate $0.002$ and locality regularization to preserve GAN expressivity. 

\subsection{Main Results}\label{results}

\begin{table}[t]
\centering
\tabcolsep 3pt
\caption{\textbf{Ablation study} on the key design components. Please see the corresponding qualitative analysis in Figure~\ref{fig:ablation}.}
\label{tab:flame_refinement_ablation}
\small
\vspace{-3mm}
\begin{tabular}{l|ccc}
\toprule
\multirow{1}{*}{\textbf{Setting}}
& PSNR $\uparrow$ & SSIM $\uparrow$ & LPIPS $\downarrow$ \\
\midrule
SuperHead (ours) & 22.15 & 0.841 & 0.243  \\
\hspace{0.5em} (a) w/o multi-view inversion & 19.86 & 0.812 & 0.283  \\
\hspace{0.5em} (b) w/o multi-expr. inversion & 21.18 & 0.824 & 0.269  \\
\hspace{0.5em} (c) w/o depth supervision & 21.74 & 0.829 & 0.251  \\
\hspace{0.5em} (d) w/o FLAME refinement & 22.03 & 0.838 & 0.246  \\
\bottomrule
\end{tabular}
\vspace{-3mm}
%
\end{table}


\begin{figure*}[t]
    \centering
    \includegraphics[width=1\linewidth]{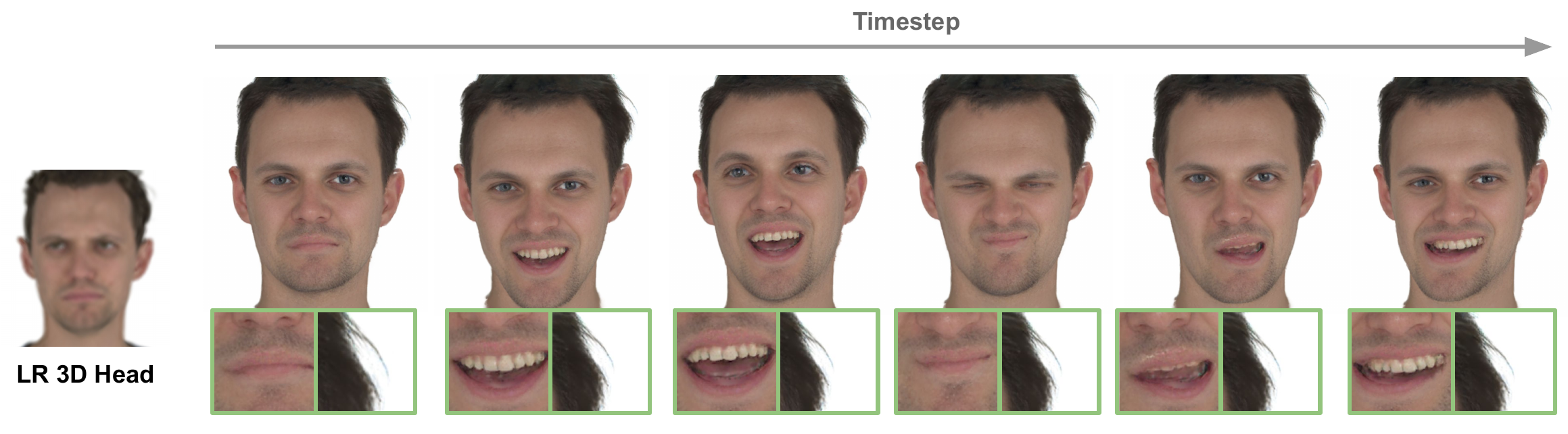}
    \vspace{-8mm}
    \caption{\textbf{Sequential frames of \ourmethod}. \ourmethod recovers fine details and facial components with multi-view/frame consistency.}
    \vspace{-3mm}
    \label{fig:sequential}
\end{figure*}

\begin{figure}[t]
    \centering
    \includegraphics[width=1\linewidth]{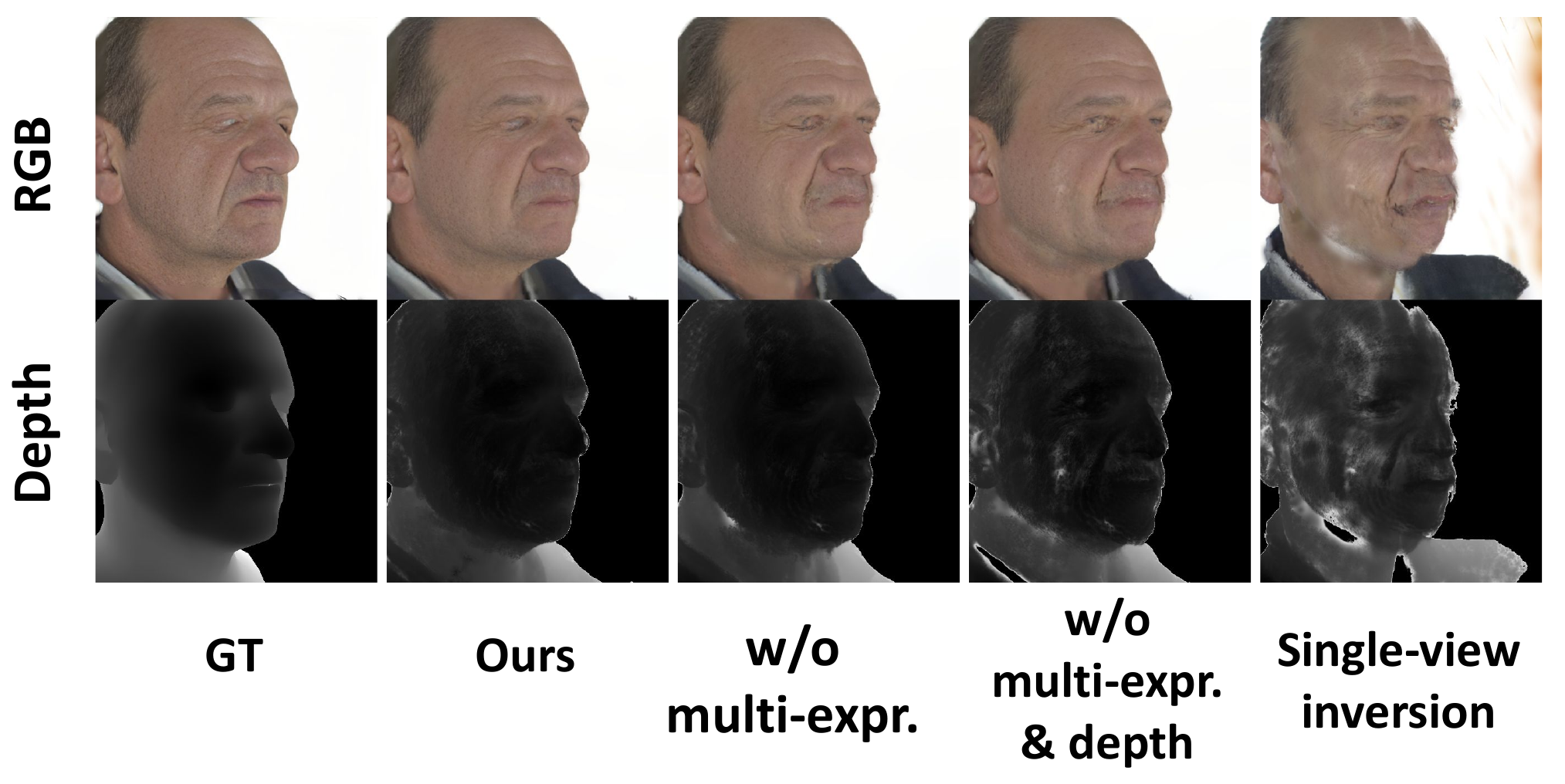}
    \vspace{-8mm}
    \caption{\textbf{Ablation studies.} We iteratively remove various design components, including multi-expression inversion, depth supervision, and multi-view inversion, and show the rendered image and depth maps of the output head avatar with novel expression and novel camera view. The results get progressively worse. 
    }
    \vspace{-3mm}
    \label{fig:ablation}
\end{figure}

We show quantitative results in Table \ref{tab:main_results}. Our method outperforms all other baselines on both multi-view and monocular video datasets. We also report inference time for each enhancement method, and \ourmethod shows great inference speed while achieving the best performance. Figure~\ref{fig:sota-compare} and Figure~\ref{fig:insta_qualitative} present qualitative comparisons between \ourmethod and the baseline approaches. Other methods exhibit blurry textures and noticeable artifacts, likely due to inconsistencies arising from independently up-scaling multi-view inputs. In contrast, \ourmethod synthesizes sharp textures and much richer facial details, e.g., facial hair and teeth. In some cases, the visual quality achieved by \ourmethod is on par with and even surpass the pseudo ground-truth (GT) HR avatar in the synthesis of certain high-frequency details, e.g., eyebrows, underscoring the effectiveness of our approach in leveraging powerful generative 3D priors for the super-resolution task. We also show rendering sequence of \ourmethod in Figure~\ref{fig:sequential}, indicating that our method can recover high-quality and consistent facial details.

\subsection{Ablation and Analysis}\label{ablation}
We conduct ablation studies to evaluate key design components and report results in Table~\ref{tab:flame_refinement_ablation} and Figure~\ref{fig:ablation}. 

\mypara{Multi-view inversion} (cf.~Section~\ref{multi_view_3d_inversion}). 
When performing 3D GAN inversion with only single-view, the quality of novel views is poor with distorted geometry and appearance (Table~\ref{tab:flame_refinement_ablation}-(a)).
In contrast, with all components enabled, \ourmethod produces high-fidelity reconstructions with smooth and coherent surfaces, demonstrating the complementary benefits of each design choice.

\mypara{Multi-expression inversion and depth supervision} (cf.~Section~\ref{dynamics_aware_inversion}). With only one expression, the 3D head often shows floating artifacts on novel expressions (Table~\ref{tab:flame_refinement_ablation}-(b)). Similarly, omitting depth supervision can result in surfaces with holes (Table~\ref{tab:flame_refinement_ablation}-(c)).

\begin{figure}[t]
    \centering
    \includegraphics[width=1\linewidth]{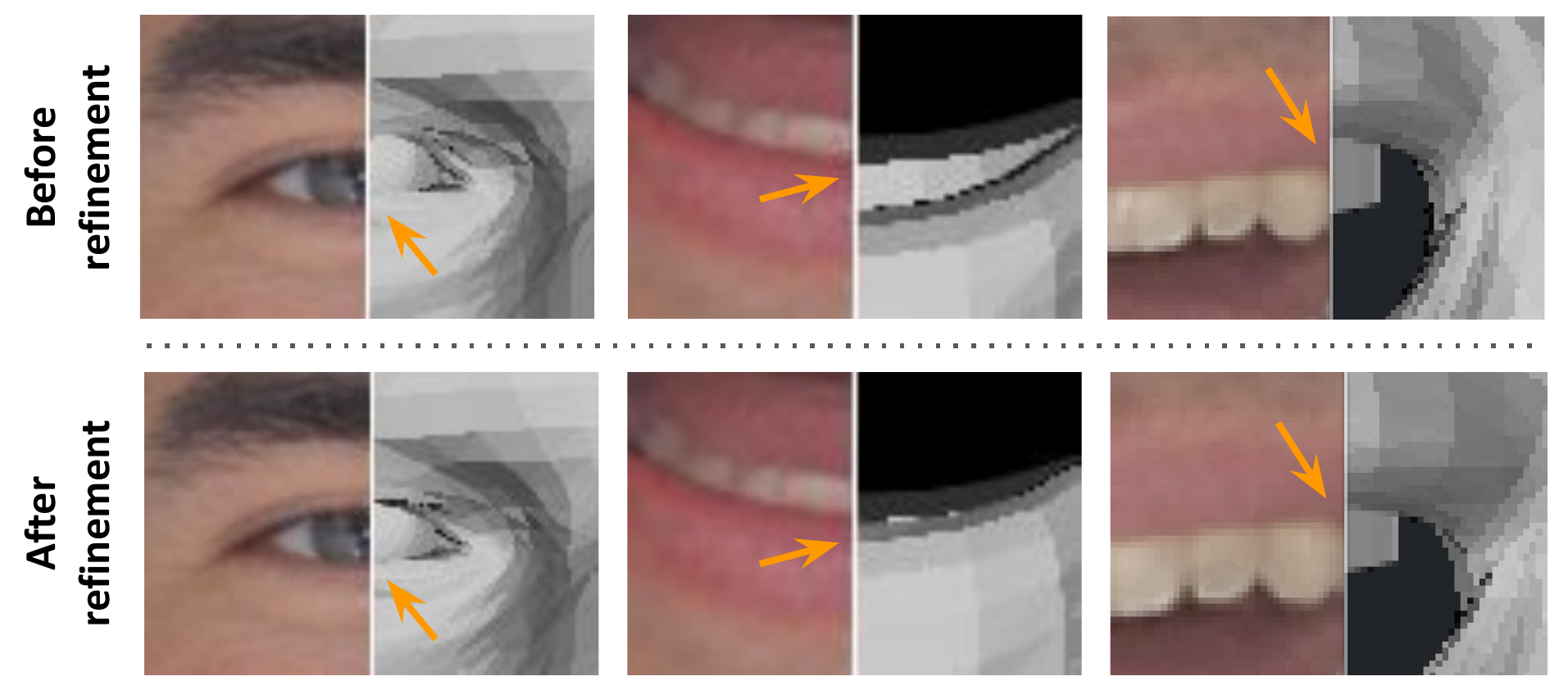}
    \vspace{-5mm}
    \caption{\textbf{Importance of geometry refinement.} Without refinement, anchor images misalign with the underlying geometry, such as the eye sockets, lower lip, and teeth. After refinement, the FLAME mesh aligns accurately with the anchor images. 
    }
    \vspace{-2mm}
    \label{fig:flame_refinement_ablation}
\end{figure}

\mypara{Geometry refinement.} 
Before 3D Gaussian binding, we refine the mesh (cf.~Section~\ref{3d_gaussian_rigging}) to ensure Gaussians are rigged to the correct parts, as shown in Figure~\ref{fig:flame_refinement_ablation}. This step also improves performance, as shown in Table~\ref{tab:flame_refinement_ablation}-(d).

\mypara{More results.}
We include videos, analysis on anchor image sampling, and results of comparison to other 3D avatar model (e.g., SplattingAvatar~\cite{shao2024splattingavatar}) in the supplementary.

\section{Discussion and Conclusion}

We present \ourmethod, a new framework for upsampling low-resolution animatable 3D head avatars. By leveraging upsampled multi-view images and depth cues across diverse expressions, \ourmethod exploits 3D GAN priors to synthesize realistic 3D Gaussian heads with geometric consistency and temporal coherence. Extensive experiments show that \ourmethod outperforms existing baselines in visual quality. We hope our method opens new possibilities for digital humans in VR, telepresence, and entertainment.

\mypara{Limitation and future work.} Our method still has challenges: it cannot handle dynamics of hair in head-shaking motion or recover $360$-degree 3D head. Please refer to the supplementary material for further discussion.

{
    \small
    \bibliographystyle{ieeenat_fullname}
    \bibliography{main}
}

\clearpage
\setcounter{section}{0}
\maketitlesupplementary

\renewcommand\thesection{\Alph{section}}
\renewcommand{\thetable}{S\arabic{table}}  
\renewcommand{\thefigure}{S\arabic{figure}}

In this supplementary material, we provide additional details and results omitted in the main text.

\section{Contribution and Limitations}
 
\textbf{Main contribution}. While many previous works have explored super-resolution (SR) in 2D content, e.g., images, or static 3D representation, e.g., 3D Gaussian, super-resolution in dynamic 3D representation remains an unexplored direction. The main challenge lies in the fact that 2D SR not only struggles with multi-view but also temporal inconsistencies, when up-sampling a dynamic 3D  representation. Our method addresses this challenge by performing multi-view and multi-expression 3D GAN inversion, ensuring that the synthesized 3D head preserves high-frequency details even when the up-sampled anchor images are inconsistent. To the best of our knowledge, this is the first attempt at super-resolution of dynamic 3D avatar representation. 

\begin{figure}[h!]
    \centering
    \includegraphics[width=1\linewidth]{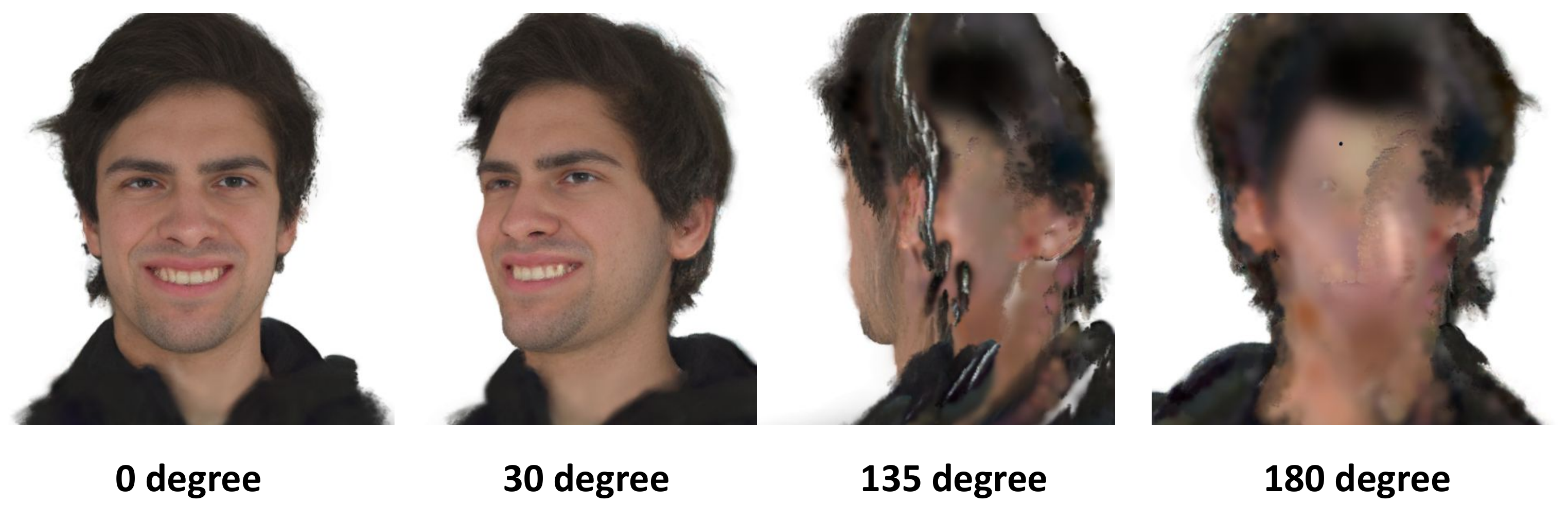}
    \vspace{-8mm}
    \caption{3D GAN struggles to synthesize the back of a human head. We rotate a synthesized head before camera to show quality gap between views of frontal and back of a head.
    }
    \vspace{-3mm}
    \label{fig:limitation}
\end{figure}

\mypara{Limitation}. The major limitation is that 3D GAN cannot synthesize complete 3D head, i.e., it struggles to generate back of a human head. The main reason is that 3D GAN is trained on FFHQ~\cite{karras2019style}, which consists only of frontal human faces. Building a large-scale human face dataset that includes views of the back of the head is a possible way to extend 3D GAN's ability of synthesizing back views of human heads. As shown in Figure~\ref{fig:limitation}, while GSGAN~\cite{hyun2024gsgan} can synthesize frontal views of high-fidelity details, it struggles to synthesize the back of the human head.

\section{Additional Implementation Details}

We adopt GSGAN~\cite{hyun2024gsgan} as our 3D GAN backbone. To make 3D GAN robust to side views of a 3D head and hairstyles, we processed FFHQ~\cite{karras2019style} by cropping the image source to include full head in the image. Then, we fine-tuned the GSGAN checkpoint on the re-cropped FFHQ dataset. Figure~\ref{fig:sota-compare-supp} shows that the fine-tuned 3D GAN can not only synthesize finer details on facial parts but also accurate hairstyles. All of our experiments, including the 3D GAN fine-tuning, were conducted on a RTX A6000 GPU. 

\section{Additional Results and Analyses}

\textbf{Spatio-temporal Quality and Identity Preservation}. While the main paper focuses on comparing static per-frame quality, we provide an additional evaluation of the spatio-temporal coherence and identity fidelity of the synthesized video sequences here. To quantify the distributional similarity between the generated and ground-truth video motion, we employ the Fréchet Video Distance (FVD) \cite{unterthiner2019accurategenerativemodelsvideo}, which utilizes an I3D backbone to extract spatio-temporal features. Furthermore, we adopt DOVER \cite{wu2023exploringvideoqualityassessment}, a learning-based blind video quality assessment (BVQA) metric, to assess perceptual quality in alignment with human aesthetic judgment. Finally, we quantify identity preservation by calculating the Cosine Similarity (CSIM) of facial embeddings extracted via a pre-trained ArcFace \cite{Deng_2022} model.

\begin{table}[h]
\centering
\tabcolsep 3pt
\caption{SuperHead outperforms all other baselines on metrics of spatio-temporal quality (FVD$\downarrow$, DOVER$\uparrow$) with high identity preservation (CSIM$\uparrow$).}
\label{tab:temporal_compare}
\small
\vspace{-3mm}
\begin{tabular}{l|ccc}
\toprule
\multirow{1}{*}{\textbf{Method}}
& CSIM $\uparrow$ & FVD $\downarrow$ & DOVER $\uparrow$\\
\midrule
GaussianAvatars (LR)~\cite{qian2024gaussianavatars} & \cellcolor{red!25}0.922 & \cellcolor{orange!25}282.48 & 15.46  \\
Video-based SR ~\cite{feng2024kalman} & 0.775 & 437.16 & 42.63  \\
SuperGaussian~\cite{shen2024supergaussian} & \cellcolor{yellow!25}0.807 & \cellcolor{yellow!25}293.59 & \cellcolor{yellow!25}59.12  \\
SR + GPAvatar~\cite{chu2024gpavatar} & 0.633 & 788.10 & \cellcolor{orange!25}74.65  \\
\textbf{SuperHead (ours)} & \cellcolor{orange!25}0.867 & \cellcolor{red!25}181.13 & \cellcolor{red!25}82.21  \\

\bottomrule
\end{tabular}
\vspace{-3mm}
%
\end{table}

As shown in Table~\ref{tab:temporal_compare}, our method achieves the best performance in temporal quality metrics, significantly reducing flickering and motion artifacts. Notably, while our method maintains high identity consistency, GaussianAvatars (LR) exhibits a slightly higher CSIM score. This is attributed to the inherent noise-invariance of face recognition models like ArcFace, which are trained to extract robust geometric signatures regardless of degradations. Since GaussianAvatars (LR) is directly optimized for down-sampled ground truth, it preserves the global structural layout (biometric signature) while our method prioritized the synthesis of high-fidelity textures, which can introduce minor, perceptually superior variations that the embedding space interprets as a slight identity shift.

\begin{figure*}[t]
    \centering
    \includegraphics[width=0.98\linewidth]{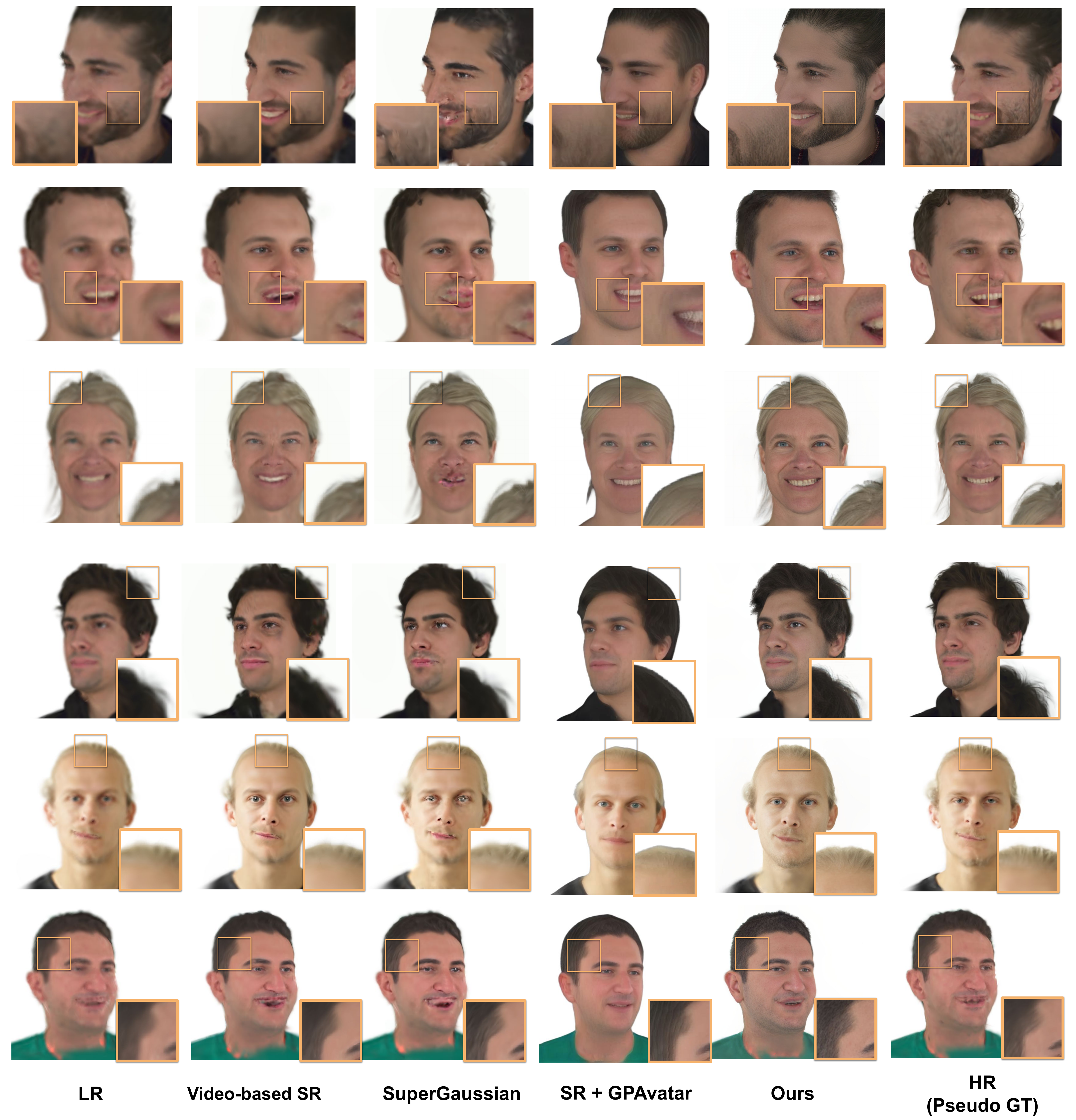}
    \vspace{-2mm}
    \caption{Additional qualitative results on the NeRSemble dataset~\cite{kirschstein2023nersemble} and INSTA~\cite{zielonka2023instant}. In addition to zooming in facial parts of results, we also show the holistic view of upsampled 3D avatar, indicating that our method can not only enhance facial expressions but also details such as hair strands. Please zoom in to check details.
    }
    \vspace{-3mm}
    \label{fig:sota-compare-supp}
\end{figure*}

\mypara{Additional qualitative results}. We show additional visual comparison of various baselines introduced in the main paper in Figure~\ref{fig:sota-compare-supp}. Our method demonstrates superior capability in recovering detailed facial expressions, e.g., corner of the mouth, but also accurate geometry of the hair. 

\begin{table}[h]
\centering
\tabcolsep 3pt
\caption{SuperHead achieves identical performance when applying to SplattingAvatar~\cite{shao2024splattingavatar} on INSTA dataset~\cite{zielonka2023instant}, proving SuperHead's generalizability to enhance diverse 3D avatar models.}
\label{tab:splattingavatar_compare}
\small
\vspace{-3mm}
\begin{tabular}{l|ccc}
\toprule
\multirow{1}{*}{\textbf{Setting}}
& PSNR $\uparrow$ & SSIM $\uparrow$ & LPIPS $\downarrow$ \\
\midrule
SplattingAvatar (LR)~\cite{shao2024splattingavatar} & 19.24 & 0.825 & 0.251  \\
SuperHead + SplattingAvatar~\cite{shao2024splattingavatar} & 23.04 & 0.834 & 0.167  \\
SuperHead + GaussianAvatars~\cite{qian2024gaussianavatars} & 23.76 & 0.864 & 0.135  \\
\bottomrule
\end{tabular}
\vspace{-3mm}
%
\end{table}

\begin{figure}[h!]
    \centering
    \includegraphics[width=1\linewidth]{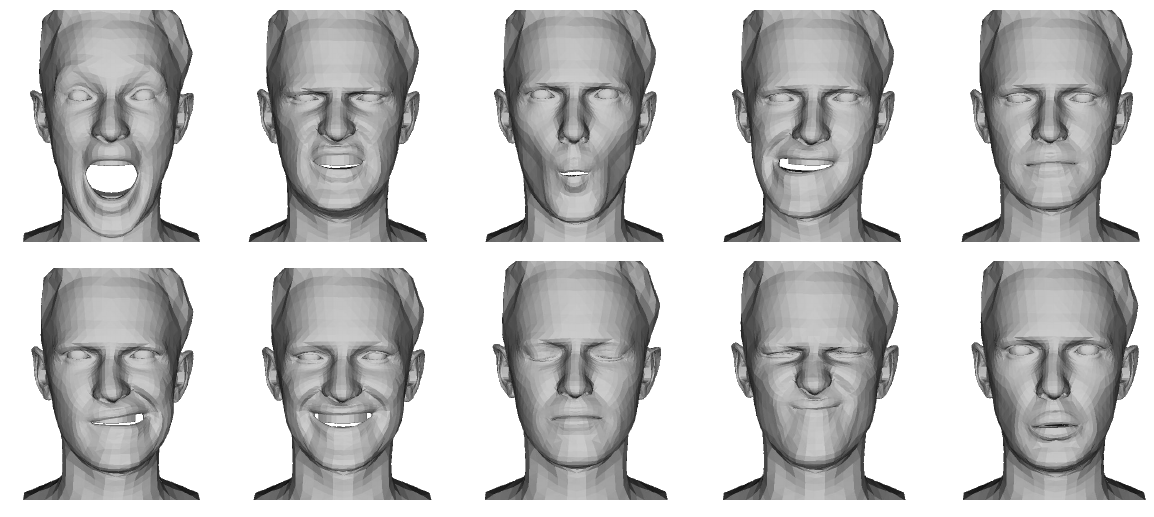}
    \vspace{-8mm}
    \caption{Expressions we used to sample anchor images.
    }
    \vspace{-3mm}
    \label{fig:expression_pool}
\end{figure}

\mypara{Comparability to other 3D avatar
model}. 
We further evaluate our method on an alternative 3D avatar model to demonstrate its generalizability. Specifically, we adopt SplattingAvatar~\cite{shao2024splattingavatar}, which, similar to GaussianAvatars~\cite{qian2024gaussianavatars}, rigs 3D Gaussians onto the FLAME mesh with a learnable normal offset to the surface. The upsampling procedure follows the same pipeline as with GaussianAvatars: we first sample and enhance anchor images from a SplattingAvatar trained on low-resolution captures, and then perform multi-view inversion along with dynamics-aware 3D refinement to optimize a 3D Gaussian head rigged on the underlying FLAME mesh. The results are reported in Table~\ref{tab:splattingavatar_compare}. We compare SplattingAvatar (LR), a 3D head model trained on low-resolution captures, with SuperHead + SplattingAvatar, the corresponding upsampled 3D head model. We show that our method successfully enhances low-quality 3D head models across different design choices, thereby demonstrating its strong generalizability.

\mypara{Anchor image sampling} As mentioned in Section 4.3 of the main paper, we perform dynamics-aware 3D GAN refinement to improve the synthesized 3D head under different expressions and motions. For this purpose, we carefully select a set of expressions to form an "expression pool", from which we sample anchor images with different camera poses. We found that a set of $10$ expressions is sufficient to achieve good performance. We show the expressions we use throughout our experiments in Figure~\ref{fig:expression_pool}. The selected expressions cover a range of facial motions, from screaming to smiling and eye-closing.

\section{Ethical and Societal Impacts}
Our work improves the quality of 3D head avatar reconstruction, which has potential benefits in areas such as telecommunication and digital content creation. At the same time, we are aware of possible risks, including issues of privacy, misuse for non-consensual content, and bias in representation. We emphasize the importance of developing and applying such techniques responsibly and with appropriate safeguards. Furthermore, like many generative methods, reconstruction results may contain certain biases if not carefully addressed. We believe it is important for future research and deployment of these techniques to be guided by principles of responsible AI, including fairness, transparency, and safeguards against malicious use.

\end{document}